\documentclass{article}
\usepackage{ijcai24}

\usepackage{times}
\usepackage{soul}
\usepackage{url}
\usepackage{enumitem}
\usepackage[hidelinks]{hyperref}
\usepackage[utf8]{inputenc}
\usepackage[small]{caption}
\usepackage{graphicx}
\usepackage{amsmath}
\usepackage{amsthm}
\usepackage{booktabs}
\usepackage{algorithm}
\usepackage{algorithmic}
\usepackage[switch]{lineno}
\usepackage{float}

\usepackage{booktabs}
\usepackage{longtable}

\usepackage{pgfplots}
\usepackage{pgfplotstable}
\pgfplotsset{compat=1.18}

\usepackage{xcolor}
\usepackage{caption, booktabs, tabularx, makecell}
\usepackage{multirow}
\usepackage{array}
\linenumbers

\title{Multi-Armed Bandits Meet Large Language Models }

\author{
    Author Name
    \affiliations
    Affiliation
    \emails
    email@example.com
}

\author{
Djallel Bouneffouf$^1$
\and
Raphael Feraud$^2$
\affiliations
$^1$ IBM Research, NewYork, USA \\
$^2$Orange Lab, Lannion, France\\
\emails
Djallel.bouneffouf@ibm.com,
raphael.feraud@orange.com
}
\begin{document}
\nolinenumbers

\maketitle

\begin{abstract}
Bandit algorithms and Large Language Models (LLMs) have emerged as powerful tools in artificial intelligence, each addressing distinct yet complementary challenges in decision-making and natural language processing. This survey explores the synergistic potential between these two fields, highlighting how bandit algorithms can enhance the performance of LLMs and how LLMs, in turn, can provide novel insights for improving bandit-based decision-making. We first examine the role of bandit algorithms in optimizing LLM fine-tuning, prompt engineering, and adaptive response generation, focusing on their ability to balance exploration and exploitation in large-scale learning tasks. Subsequently, we explore how LLMs can augment bandit algorithms through advanced contextual understanding, dynamic adaptation, and improved policy selection using natural language reasoning. By providing a comprehensive review of existing research and identifying key challenges and opportunities, this survey aims to bridge the gap between bandit algorithms and LLMs, paving the way for innovative applications and interdisciplinary research in AI.
\end{abstract}

\section{Introduction}
%\subsubsection{Motivation for Combining Bandit Algorithms and Large Language Models (LLMs)}
LLMS such as GPT \cite {Brown2020}, BERT \cite {Delvin2019}, and T5 \cite {Raffel2020} have revolutionized natural language processing by excelling in tasks like text generation, summarization, and dialogue systems. However, optimizing LLMs for specific applications often involves challenges like balancing exploration (testing novel strategies) and exploitation (leveraging learned strategies). Bandit algorithms \cite {bouneffouf2020survey,bouneffouf2023multi}, designed to address such trade-offs, provide a powerful framework for optimizing decision-making under uncertainty. By integrating bandit approaches into the training, tuning, and application of LLMs, researchers can achieve dynamic adaptability, efficiency, and enhanced performance.

Conversely, LLMs possess the ability to process and generate human-like text, enabling them to provide contextual insights and reasoning. This capability can enhance bandit algorithms by incorporating richer context, natural language understanding, and user feedback into the decision-making process. The interplay between these two fields opens the door to innovative AI applications that are both adaptive and contextually aware.

%\subsubsection{Importance of Exploring the Interaction Between Bandits and LLMs}
While Bandit algorithms and LLMs have independently demonstrated success across various domains, the exploration of their intersection remains underdeveloped. By investigating how these paradigms complement each other, researchers can unlock new possibilities, such as improving LLM performance in personalized recommendations or utilizing LLMs to address the contextual limitations of bandit-based decision systems. This exploration not only advances the theoretical understanding of both fields but also provides practical solutions for complex real-world problems, including adaptive dialogue systems, dynamic content generation, and human-AI collaboration.

%\subsubsection{Contributions}
This survey aims to provide a comprehensive overview of the emerging synergy between Bandit algorithms and LLMs \cite{banditllm24}. It explores:
\begin {itemize}
\item How Bandit algorithms can enhance the efficiency and adaptability of LLMs,
\item How LLMs can contribute to the contextual and adaptive decision-making capabilities of Bandit algorithms,
\item Applications where the integration of these paradigms offers tangible benefits,
%Open challenges and future research directions at the intersection of these fields.
\item  Discussion of open problems, research gaps, and potential future directions in combining Bandit algorithms and LLMs.
\end {itemize}
By synthesizing the current state of research and offering a roadmap for future exploration, this paper aims to serve as a foundational resource for researchers and practitioners interested in the intersection of Bandit algorithms and LLMs.

\section{Leveraging Bandit to Enhance LLM}

The integration of Bandit algorithms into machine learning and natural language processing has been an active area of research. While traditional applications of Bandit algorithms include many domains such as online recommendation systems \cite {bouneffouf2023question,BouneffoufRCF17,BouneffoufLUFA14,IdeMBA25,ChenLG024}, healthcare \cite {LinB22,ijcnnLinBC22,Lin0C22,LinC0RR20}, automated machine learning \cite {dj2021,0001C21,0001RV0BSW0G20,RiemerKBF19}, question answering \cite{FUKR21}, telecommunication \cite {Zafar2023,Delande2021}, material science \cite {Kishimoto00RRWP22}, Reinforcement Learning \cite {lin2023psychotherapy,epperlein2022reinforcement,BalakrishnanBMR19,NoothigattuBMCM19}, recent works have explored their role in optimizing LLMs. This section reviews key contributions in the following areas: fine-tuning and training, prompt optimization, adaptive response generation, and evaluation strategies.  

\subsection{Fine-Tuning and Training}
Fine-tuning large-scale language models is a resource-intensive process that requires selecting the most informative data samples, optimizing hyperparameters, and ensuring generalization while keeping computational costs manageable. Traditional methods often rely on heuristics or grid search techniques, but recent advances have demonstrated that multi-armed bandits (MABs) and contextual bandits can enhance various aspects of the fine-tuning process.

\subsubsection{Active Learning for Data Selection}
One of the critical challenges in fine-tuning LLMs is determining which data samples contribute most for improving the model. Instead of training on all available data, which can be computationally expensive and redundant, Bandit algorithms have been employed for active learning to prioritize the most informative data points.

Uncertainty-Based Sampling: Bandit models help identify samples where the model is most uncertain, selecting them for fine-tuning to improve performance in underrepresented areas \cite{settles09}.

Diversity-Promoting Selection: Contextual bandits can be used to balance between selecting uncertain samples and ensuring diverse coverage of the data distribution, preventing biases in model adaptation\cite{xia2024llm,chang2023learning,1011453627821,bouneffouf2014contextual,wahed2023marble}.

Cost-Efficient Learning: By dynamically allocating resources to the most impactful data points, Bandit-based active learning reduces the amount of labeled data required while maintaining or improving model performance \cite{li2025llm}.
\subsubsection{Hyperparameter Tuning}
Fine-tuning performance is highly dependent on selecting optimal hyperparameters such as learning rates, batch sizes, and dropout rates. Traditional methods like grid search or random search can be inefficient, whereas Bandit-based optimization methods,  provide a more adaptive approach.

Bayesian Optimization for Hyperparameters: Framed as a bandit problem, Bayesian optimization iteratively refines hyperparameter choices by exploring promising configurations while minimizing unnecessary trials \cite{li2018hyperband,snoek2012practical}.
Multi-Fidelity Optimization: Some Bandit approaches allocate resources dynamically to hyperparameter settings that show early promise, reducing wasteful full-scale evaluations \cite{mulakala2024adaptive}.
%Automated Learning Rate Schedules: Adaptive exploration strategies using Bandits help in dynamically adjusting learning rates to balance stability and convergence speed.
\subsubsection{Adaptive Gradient-Based Methods}
Beyond data selection and hyperparameters, Bandit algorithms have also been applied to optimize the learning process itself by improving gradient-based optimization techniques.

Thompson Sampling for Gradient Updates: Bandit-inspired approaches such as Thompson Sampling adaptively adjust gradient update rules, improving convergence rates and reducing overfitting \cite{liu2020adam}.
Exploration in Optimization Strategies: By dynamically selecting among different optimization techniques (e.g., Adam, RMSProp, SGD), Bandit models can guide learning toward more efficient convergence patterns \cite{song2024trial}.
Adaptive Batch Size Selection: Bandit algorithms have been proposed to adjust batch sizes in real time, optimizing trade-offs between convergence speed and computational efficiency \cite{lisicki2023bandit}.

\subsubsection{Impact on Fine-Tuning Efficiency}
The integration of Bandit-based decision-making into fine-tuning strategies has demonstrated significant improvements in both data efficiency and training performance for LLMs. By leveraging exploration-exploitation principles, these approaches lead to faster convergence with reduced training time, improved model generalization through selective data exposure, and more efficient allocation of computational resources.
Future research can explore for instance, the selection of LLM experts for reducing the query cost, or the use of Bandit-driven curriculum learning to further enhance the adaptability and robustness of LLM training\cite{lisicki2023bandit}.

\subsection{Prompt Optimization}
Prompt engineering plays a crucial role in determining the quality, relevance, and coherence of responses generated by large language models (LLMs). Manually crafting optimal prompts requires domain expertise and iterative experimentation, making it a time-consuming process. Bandit algorithms provide a systematic way to dynamically optimize prompts by continuously exploring different formulations and selecting those that yield the best performance based on predefined reward signals.

\subsubsection{Multi-Armed Bandit for Dynamic Prompt Selection}

%One of the primary challenges in prompt optimization is identifying the most effective phrasing for a given task. 
Recent research has formulated prompt selection as a multi-armed bandit (MAB) problem, where different prompt variants represent different "arms," and the LLM’s response quality serves as the reward signal.

Exploration vs. Exploitation: Instead of relying solely on predefined prompts, Bandit-based approaches allow for continuous exploration of new formulations while exploiting the most successful ones. This adaptive strategy helps identify prompts that produce higher precision, coherence, and information \cite{shi2024efficient}.

Automated Prompt Discovery: MAB algorithms can iteratively refine prompts by adjusting keywords, phrasing, or sentence structure, optimizing for factors such as fluency, factual consistency, and response diversity \cite{gao2025prompt}.

Evaluation Metrics as Rewards: Bandit models can use various reward functions based on response quality metrics, including BLEU scores (for linguistic similarity), factual consistency (measured via external verification models), or human feedback ratings\cite{grams2025disentangling}.

\subsubsection{Chain-of-Thoughts}
To enhance the ability of large language models (LLMs) to tackle complex reasoning problems, chain-of-thought (CoT) methods have been introduced to guide LLMs through step-by-step reasoning, enabling problem-solving from simple to complex tasks \cite {Wei2022}. State-of-the-art approaches to generate these reasoning chains are based on interactive collaboration, where the learner produces intermediate candidate thoughts that are evaluated by the LLM, influencing the generation of subsequent steps .

However, a significant, yet underexplored challenge is that LLM-generated evaluations are often noisy and unreliable, which can mislead the selection of promising intermediate thoughts. To address this issue, %inspired by Vapnik’s principle, 
the authors in \cite{zhang2024generating} propose a bandit-based pairwise comparison framework instead of conventional point-wise scoring. In each iteration, the intermediate thoughts are randomly paired, and the LLM is directly prompted to select the most promising option from each pair. This iterative comparison process, framed as a dueling bandit problem \cite {Urvoy2013}, allows for adaptive exploration and exploitation of promising reasoning paths while reducing the impact of noisy feedback.

\subsubsection{Contextual Bandits for Personalized Prompting}

Beyond general prompt selection, optimizing prompts for specific users or contexts is essential for improving interaction quality. Contextual bandits extend traditional MAB models by incorporating contextual features such as user intent, domain-specific requirements, and interaction history.

Adaptive Prompt Tuning: Contextual bandits dynamically adjust prompt attributes like length, specificity, and style based on the user's query type and preferences \cite{lau2024personalized,chen2024online}.

Personalized User Interaction: By analyzing past interactions, contextual bandit models can tailor prompts to individual users, ensuring more relevant and engaging responses \cite{dai2025multi}.

Domain-Specific Adaptation: In specialized fields such as legal or medical AI applications, contextual bandits can fine-tune prompts to align with domain-specific jargon and information retrieval needs \cite{upadhyay2019banditapproachposteriordialog}.

%\subsubsection{Reinforcement Learning with Human Feedback (RLHF) and Bandit-Based Exploration}

%Reinforcement Learning with Human Feedback (RLHF) is widely used for fine-tuning LLMs to align their outputs with human preferences. While RLHF typically relies on policy optimization techniques, incorporating Bandit-based exploration strategies can enhance its efficiency.

%Reducing Human Labeling Effort: By using Bandit algorithms to prioritize human labeling for the most uncertain or impactful prompts, RLHF can be made more data-efficient \cite{chang2023learning,1011453627821,bouneffouf2014contextual,wahed2023marble}.

%Optimizing Reward Model Updates: Bandit-inspired strategies help in dynamically adjusting the weight given to different types of feedback (e.g., explicit user ratings vs. implicit engagement signals).

%Balancing Exploration and Exploitation: Traditional RLHF may overfit to specific types of human feedback, but Bandit-based approaches ensure a better balance between learning from past feedback and exploring new prompt formulations.

\subsubsection{Impact on Prompt Engineering Efficiency}

By leveraging Bandit algorithms, LLMs can be automatically guided toward more effective and adaptive prompt formulations, reducing reliance on manual prompt engineering. The key benefits of Bandit-based prompt optimization include:

- improved Response Quality, continuous refinement leads to prompts that generate more accurate and contextually relevant responses,
- reduced Trial-and-Error Costs, automated exploration of prompt variations decreases the need for extensive manual experimentation,
- scalability across domains, Bandits can adapt prompts for various applications, from chatbots and virtual assistants to domain-specific AI systems.
Future research directions could explore integrating hierarchical Bandit models for multi-step prompt optimization, 
%leveraging large-scale reinforcement learning techniques, 
and developing hybrid approaches that combine rule-based heuristics with adaptive learning strategies.

\subsection{Adaptive Response Generation}
Generating high-quality responses requires balancing creativity and relevance, especially in conversational AI and dialogue systems. Bandit-based approaches have been used to dynamically adjust generation strategies based on user interactions and feedback.  

Exploration-Exploitation Trade-Off in Response Selection: Thompson Sampling and Upper Confidence Bound (UCB) algorithms have been employed to explore diverse response strategies while gradually shifting towards more rewarding (i.e., coherent, engaging) responses \cite{xia2024llm}.
Adaptive Sampling for Diversity: Multi-armed bandits have been used to balance novelty vs. coherence, ensuring that generated responses are neither too predictable nor too random \cite{hoveyda2024aqaadaptivequestionanswering}.
Conversational Personalization: Bandit-driven methods allow for adaptive dialogue generation by continuously learning from user interactions and refining response styles accordingly \cite{cai2021bandit}.

These studies indicate that Bandit algorithms can improve LLM-generated responses by continuously adapting to changing user preferences and real-time feedback.  

\subsection{Evaluation Strategies}
Evaluating the quality of outputs generated by large language models (LLMs) presents significant challenges, as assessment criteria such as fluency, coherence, factual accuracy, and user preference can be highly subjective. Traditional evaluation methods often rely on human annotations, rule-based metrics, or pre-trained scoring models, all of which have limitations in scalability, consistency, and adaptability. Bandit-based approaches provide a promising solution by dynamically optimizing evaluation strategies, reducing human annotation effort while improving the quality and efficiency of feedback collection \cite{xia2024convergence}.

Reinforcement Learning with Human Feedback (RLHF): the human preferences guide model optimization \cite{kaufmann2023survey}. While RLHF typically relies on policy optimization techniques, incorporating Bandit-based exploration strategies can enhance its efficiency: traditional
RLHF may overfit to specific types of human feedback, but Bandit-based approaches ensure a better balance between learning from past feedbacks and exploring new feedbacks.

Optimizing Reward Model Updates: Bandit-inspired
strategies help in dynamically adjusting the weight given to different types of feedback (e.g., explicit user ratings vs. implicit engagement signals).

%Multi-Armed Bandit for Reward Models:
%Reinforcement Learning with Human Feedback (RLHF) is a crucial component in fine-tuning LLMs, where human preferences guide model optimization \cite{kaufmann2023survey}. 

Dynamic Reward Adjustment: As models improve over time, bandit algorithms dynamically update evaluation criteria to focus on emerging weaknesses, ensuring that reinforcement learning continues to drive meaningful improvements \cite{yang2024rewards}.

Adapting Metrics Based on Task-Specific Performance: Different tasks (e.g., summarization, translation, open-ended generation) require different evaluation criteria. Bandit algorithms adaptively select the most relevant metrics for each task \cite{xia2024llm}.

Continuous Optimization of Evaluation Pipelines: Instead of relying on static metric weighting, bandit algorithms iteratively adjust how much weight is assigned to each metric based on real-world feedback \cite{wu2024streambench}.

\subsection{Summary}

\begin{table*}[h!]
    \centering
    \begin{tabular}{l|l|l|l}
        \toprule
        \textbf{Aspect} & \textbf{Challenges} & \textbf{Bandit Solutions} & \textbf{Impact} \\
        \midrule
        Training & High cost, inefficiency & Adaptive sampling, tuning & Faster, cheaper training \\
        Prompt Optimization & Manual tuning is slow & Contextual, dueling bandits & Better responses \\
         Personalization & User preference variation & Contextual bandits, fine-tuning & Tailored, user-centric responses \\
        RLHF & Overfitting, annotation cost & Optimized feedback selection & Fair, efficient learning \\
        Future & Scalability, robustness & Hierarchical, hybrid models & Smarter, adaptable LLMs \\
        \bottomrule
    \end{tabular}
    \caption{Bandit Algorithms for LLM Improvement}
    \label{tab:bandits}
\end{table*}

The Key Takeaways from Table \ref {tab:bandits}:

The integration of Bandit algorithms into Large Language Models (LLMs) has demonstrated significant advancements across multiple aspects of model optimization, including fine-tuning, prompt engineering, adaptive response generation, and evaluation strategies.

Fine-Tuning and Training: Traditional fine-tuning methods rely on exhaustive data selection and hyperparameter tuning, often leading to inefficiencies. Bandit algorithms, particularly Multi-Armed Bandits (MABs) and contextual bandits, offer an adaptive alternative by prioritizing data samples based on uncertainty and diversity, optimizing hyperparameters dynamically, and refining gradient-based methods. These approaches improve model generalization, reduce computational costs, and accelerate convergence.

Prompt Optimization: Prompt engineering plays a critical role in LLM performance, yet manual tuning is labor-intensive. Bandit-based strategies, such as dynamic prompt selection, enable continuous exploration of different prompt formulations while exploiting the most effective ones. Contextual bandits further personalize prompt adaptation based on user preferences and domain-specific requirements. Additionally, a dueling bandit framework mitigates noisy LLM-generated evaluations in chain-of-thought reasoning.

The increasing focus on personalization: the ability to tailor responses and strategies based on individual user preferences. Bandit algorithms, particularly contextual bandits, are effective tools for enabling this personalization by dynamically adjusting to different user needs and contexts. This customization enhances user experiences and optimizes interactions, offering more relevant and targeted outputs.

Reinforcement Learning with Human Feedback (RLHF): Bandit algorithms enhance RLHF by optimizing reward model updates, prioritizing human annotations, and balancing exploration-exploitation trade-offs. This leads to more efficient learning, reducing overfitting to specific feedback patterns.

Impact and Future Directions: Bandit-driven methods significantly improve LLM efficiency, reducing trial-and-error costs and increasing adaptability across domains. Future research could explore hierarchical Bandit models \cite{hong2022hierarchical}, hybrid rule-based approaches \cite{butz2006rule}, and Bandit-driven curriculum learning to further enhance robustness in LLM optimization \cite{graves2017automated}.

\section{Leveraging LLMs to Enhance Bandit Algorithms}

While Bandit algorithms have proven useful for optimizing LLM performance, the reverse interaction—using LLMs to improve Bandit-based decision-making—remains an emerging research frontier. This section explores key areas where LLMs can enhance Bandit algorithms.

\subsection{Contextual Understanding}
Bandit algorithms, particularly contextual bandits, rely on feature representations to make informed decisions. Traditionally, these features are manually engineered, often constrained by domain knowledge and predefined structures. However, LLMs offer a powerful alternative by automatically extracting high-dimensional semantic-rich representations from unstructured textual data.

Feature Extraction from Textual Inputs: LLMs can process raw text (e.g., user queries, product descriptions, or conversational history) and generate embeddings that encode deep contextual relationships \cite{baheri2023llms}. These embeddings can serve as input features for contextual bandits, improving their ability to distinguish between different contexts. 

Disambiguation and intention recognition \cite{kelley2012context}: In many applications, reward signals depend on understanding nuanced user intent. LLMs can classify user intentions, sentiment, and preferences from interactions, providing a more informed contextual representation for bandit decision-making.

Example of application, Personalized Recommendation Systems: In online platforms, contextual bandits are used to recommend articles, advertisements, or products. Instead of relying on static user profiles, LLMs can dynamically extract real-time user preferences from chat logs, search history, or reviews, enhancing bandit-based recommendations.
By enriching contextual bandits with deeper, more adaptive feature representations, LLMs enable more accurate and flexible decision-making in dynamic environments.

\subsection{Policy Adaptation}
Traditional Bandit algorithms adjust their policies based on numerical reward feedback, which may not always provide high-level strategic insights about changing environments. LLMs, with their ability to analyze trends, summarize past interactions, and predict future conditions, can enhance Bandit policy adaptation in several ways:

Generating Adaptive Exploration Strategies \cite{khoramnejad2025generative,zhang2021adaptive} : Instead of using fixed exploration heuristics (e.g., epsilon-greedy or UCB), LLMs can analyze historical data to dynamically suggest exploration rates based on environmental changes.

Policy Updates in Non-Stationary Environments: Many real-world applications involve evolving reward distributions (e.g., user interests shift over time) \cite{de2023llm}. LLMs can predict reward drift by analyzing sequential user behavior and adjust exploration-exploitation trade-offs accordingly.

Example Application: Dynamic Content Moderation: In social media platforms, moderation policies must continuously evolve based on emerging trends and user reports. LLMs can monitor discourse changes and recommend real-time policy updates to bandit-based content moderation systems.
By integrating LLM-driven reasoning, bandit models can proactively adapt policies instead of merely reacting to numerical rewards, leading to more robust and resilient decision-making strategies.

\subsection{Exploration-Exploitation Insights}
Balancing exploration (trying new actions) and exploitation (favoring known high-reward actions) is a core challenge in Bandit algorithms. Traditional approaches rely on statistical methods to manage this trade-off, but they often lack long-term foresight. LLMs can enhance this balance by incorporating historical insights, domain knowledge, and predictive modeling:

Forecasting Long-Term Reward Trajectories: LLMs can analyze past interactions to predict potential reward distributions for different actions, allowing Bandit models to make more informed exploration choices \cite{chen2023non,yacobi2023robust}.

Semantic Similarity for Knowledge Transfer \cite{xu2024reasoning}: In cases where limited feedback is available, LLMs can assess the semantic similarity between different arms, enabling bandit algorithms to transfer knowledge across related decisions .

Example Application – In medical treatment recommendation systems \cite{jin2024matching}, exploration must be carefully balanced with patient safety. LLMs can analyze past case studies, clinical trial results, and medical literature to predict treatment effectiveness, guiding bandit-based treatment selection.
By incorporating LLM-driven predictive modeling, Bandit algorithms can improve exploration efficiency and reduce suboptimal selections, leading to faster and more reliable convergence to optimal actions.

\subsection{Natural Language Feedback}
One of the major limitations of traditional Bandit learning is its reliance on explicit numerical reward signals, which can be sparse or difficult to obtain. In many real-world applications, user feedback is provided in natural language, requiring interpretation before it can be used to update Bandit policies. LLMs can bridge this gap by converting qualitative feedback into structured rewards:

Sentiment Analysis for Implicit Reward Extraction: Instead of relying on explicit ratings (e.g., 1–5 stars), LLMs can analyze customer reviews, chat logs, or social media comments to extract implicit satisfaction signals for Bandit learning \cite{parthasarathy2025multilinguality}.

Summarizing Feedback for Reward Calibration: LLMs can condense large volumes of user responses into structured insights, enabling Bandit algorithms to adjust their policies without requiring exhaustive manual labeling \cite{hoveyda2024aqaadaptivequestionanswering}.

Example Application – Interactive AI Assistants: In customer service chatbots, users often provide feedback in natural language (e.g., "That wasn’t helpful" or "Great answer!"). LLMs can translate these subjective responses into quantifiable rewards for Bandit-based adaptive dialogue optimization.
This capability allows Bandit algorithms to continuously learn from human interactions, improving adaptability and responsiveness without requiring structured feedback mechanisms.

\subsection{Summary}

\begin{table*}[h]
    \centering
    \begin{tabular}{l|l|l|l}
        \toprule
        \textbf{Aspect} & \textbf{Challenges} & \textbf{Bandit Solutions} & \textbf{Impact} \\
        \midrule
        Applications & Complex decision-making & LLMs enhance fairness, optimization & Improved task performance \\
        Integration & Varying LLM usage & Reward modeling, adaptive learning & Better strategy refinement \\
        Adaptability & Bias, uncertainty & LLMs improve response to changes & More robust decision-making \\
        Efficiency & High computation cost & Few studies on scalability & Limited deployment feasibility \\
        Benchmarking & Lack of standardization & Need for common evaluation metrics & Hard to assess generalizability \\
        \bottomrule
    \end{tabular}
    \caption{LLM Applications and Challenges in Bandit Problems}
    \label{tab:llm_bandits}
\end{table*}

This summary outlines some key takeaways from the previous section that explores the application of Large Language Models (LLMs) in various aspects of bandit problems. Here’s a breakdown of each point:

Diversity of Applications: LLMs are being applied in a wide range of bandit-related tasks, enhancing both basic decision-making processes (like regret minimization) and more complex objectives, such as fairness in multilingual settings. This diversity shows the potential of LLMs to improve different facets of the bandit problem.

LLM Integration: Different studies utilize LLMs in varying ways. Some focus on reward modeling, where LLMs help predict the rewards of certain actions. Others employ LLMs for adaptive learning, adjusting strategies over time based on new data. There are also cases where LLMs refine bandit strategies, optimizing decision-making in real-time.

Impact on Bandit Algorithms: LLMs appear to improve decision-making by making it more adaptable and reducing biases in dynamic environments. This highlights the ability of LLMs to enhance the performance of bandit algorithms, especially in uncertain or changing contexts.

Limited Exploration on Efficiency: While LLMs contribute to better decision-making, the studies seem to overlook their computational cost, scalability, and robustness under adversarial conditions. These are important factors when deploying LLM-enhanced bandit systems in real-world applications.

Few Benchmarks Across Studies: The comparison of different LLM-assisted methods across studies is lacking, which makes it difficult to assess the generalizability of these approaches. Standardized benchmarks would be valuable for understanding the relative performance of various techniques and guiding future research.

\section{Applications and Use Cases}

The combination of Bandit algorithms and LLMs has the potential to revolutionize various domains by enabling more adaptive, intelligent, and efficient decision-making systems. Below, we explore key applications where these technologies can be integrated.

\subsection{Personalization and Recommendation Systems}
Modern recommendation engines aim to provide users with personalized content, whether in e-commerce, media streaming, or online advertising. Traditional recommendation systems rely on collaborative filtering, reinforcement learning, or rule-based heuristics. The integration of Bandit algorithms and LLMs enhances these systems in several ways:

Dynamic Adaptation: Bandit algorithms optimize content selection by continuously learning from user feedback, ensuring that recommendations remain relevant even as user preferences evolve.
Enhanced User Understanding: LLMs improve user profiling by analyzing explicit feedback (e.g., product reviews, search queries) and implicit signals (e.g., browsing history, click patterns).
Cold-Start Problem Mitigation: In cases where new users or items enter the system, LLMs can infer preferences from textual metadata, while Bandit algorithms optimize exploration strategies to accelerate personalization \cite{ye2024lolallmassistedonlinelearning}.

\subsection{Dialogue Systems}
Conversational AI plays a crucial role in virtual assistants, customer support, and interactive chatbots \cite{hoveyda2024aqaadaptivequestionanswering}. While LLMs enable human-like text generation, they often struggle with long-term optimization and engagement strategies. By incorporating Bandit algorithms, dialogue systems can:

Optimize Response Selection: Bandits help balance response diversity and informativeness, ensuring that the AI remains engaging without becoming repetitive.
Personalize Interactions: LLMs extract user sentiment and intent, while Bandits optimize response styles and topic transitions for a more adaptive conversation.
Improve Customer Satisfaction: In customer support applications, Bandit algorithms prioritize high-value responses, continuously refining strategies based on real-time feedback.

\subsection{Autonomous Systems}
Autonomous decision-making in robotics, drones, and self-driving vehicles requires adaptability in uncertain environments. Traditional control strategies rely on pre-programmed rules or reinforcement learning models, which can struggle with real-time adjustments. By combining Bandits and LLMs, autonomous systems can:

Enhance Situation Awareness: LLMs interpret multimodal data (e.g., sensor readings, human instructions), while Bandits optimize decision-making.
Refine Action Selection: Instead of rigid control policies, Bandit algorithms adjust movement strategies based on real-time conditions, improving efficiency and safety.
Enable Human-AI Collaboration: LLMs process natural language commands, while Bandits optimize task allocation between humans and autonomous agents.

\subsection{Healthcare and Education}
Adaptive learning and personalized healthcare are two fields that benefit greatly from AI-driven optimization. The combination of Bandit algorithms and LLMs enables:

Personalized Treatment Recommendations: Bandits refine treatment plans based on patient responses, while LLMs analyze medical records and clinical notes to enhance decision-making.
Adaptive Learning Pathways: In education, Bandits determine optimal learning resources, while LLMs provide contextual explanations and engagement-driven adaptations.
Enhanced Diagnostics: LLMs extract insights from medical literature, and Bandit algorithms optimize test selection to minimize unnecessary procedures.

These applications illustrate the broad potential of combining Bandit algorithms and LLMs, paving the way for more intelligent, responsive, and human-centric AI systems.

\section{Challenges and Open Problems}

Despite the promising potential of combining Bandit algorithms with LLMs, several challenges remain to be addressed. This section outlines key open problems in this research domain.

Adaptive Reasoning Pathways: Employing bandit strategies to select among various reasoning pathways or prompts during inference could enhance LLMs' performance on complex tasks. By dynamically choosing the most promising reasoning strategies, models might achieve better accuracy and coherence in their outputs \cite{deepseekai2025deepseekr1incentivizingreasoningcapability}.

Trust and Interpretability: Ensuring transparency in decision-making processes is crucial for real-world deployment. Bandit-driven decisions influenced by LLMs could benefit from some theoretical guarantees such as regret upper bound, and should be interpretable and explainable, particularly in critical applications like healthcare and autonomous systems.

Multi-Agent Scenarios: Integrating Bandit algorithms and LLMs in multi-agent environments introduces complexity in coordination, communication, and decision-making. New approaches are required to ensure optimal cooperation and competition among multiple agents using these technologies.

Addressing these challenges is essential for advancing the integration of Bandit algorithms and LLMs into practical, scalable, and robust AI systems.

\section{Future Directions}

The intersection of Bandit algorithms and LLMs presents numerous opportunities for future research. This section outlines key directions for advancing this interdisciplinary field.

Opportunities for Interdisciplinary Research:
The fusion of Bandit algorithms and LLMs benefits from collaborations across multiple disciplines, including reinforcement learning, natural language processing, cognitive science, and human-computer interaction. By integrating insights from these fields, researchers can develop more robust and adaptive decision-making frameworks.

Emerging Trends: Recent advancements in multi-modal LLMs and hybrid algorithms open new avenues for research. Multi-modal models, which process diverse data types (e.g., text, images, audio), can enrich Bandit learning by incorporating varied contextual signals. Similarly, hybrid algorithms that combine deep learning and Bandit strategies can enhance exploration-exploitation balance in complex environments.

Potential Benchmarks for Evaluating Combined Approaches:
Establishing standardized benchmarks is essential for evaluating the effectiveness of integrating Bandit algorithms with LLMs. Future research should focus on creating datasets, simulation environments, and performance metrics that reflect real-world use cases. Benchmarking efforts will provide a foundation for comparing different methodologies and accelerating progress in the field.

Exploring these future directions will help refine the synergy between Bandit algorithms and LLMs, fostering advancements in AI-driven decision-making and adaptive learning systems.

\section{Conclusion}
The integration of Bandit algorithms and Large Language Models (LLMs) represents a promising frontier in AI-driven decision-making. This survey has explored how Bandit algorithms can enhance LLM optimization and how LLMs, in turn, can improve Bandit-based strategies through contextual understanding, policy adaptation, and natural language feedback. The synergy between these approaches enables more intelligent, adaptive, and human-aligned AI systems across various applications, including recommendation systems, dialogue agents, healthcare, and autonomous systems.

Despite these advantages, significant challenges remain, such as scalability, real-time decision-making, and interpretability. Addressing these issues will require interdisciplinary research, novel algorithmic developments, and standardized benchmarks to evaluate progress. Future work should focus on refining hybrid models, improving computational efficiency, and exploring multi-agent interactions.

By continuing to develop and integrate these techniques, researchers can unlock new opportunities for more effective and trustworthy AI systems, ultimately advancing the field of reinforcement learning and intelligent decision-making.

%\clearpage

\bibliographystyle{named}
\bibliography{ijcai24}

\end{document}